\documentclass{article}

\usepackage[a4paper]{geometry}
\usepackage[T1]{fontenc}    % use 8-bit T1 fonts
\usepackage{libertine}
\usepackage[natbib=true,style=numeric,sorting=none,url = false]{biblatex}
\usepackage{hyperref}
\usepackage{url}   % simple URL typesetting
\usepackage{booktabs}       % professional-quality tables
\usepackage{amsfonts}       % blackboard math symbols
\usepackage{amsmath}
\usepackage{nicefrac}       % compact symbols for 1/2, etc.
\usepackage{microtype}      % microtypography
\usepackage[dvipsnames]{xcolor}
\usepackage{xspace}
\usepackage{physics}
\usepackage{graphicx}
\usepackage{subcaption}
\usepackage{siunitx}
\usepackage{float}
\usepackage{circledsteps}

\definecolor{pastelred}{rgb}{1.0, 0.41, 0.38}

\newcommand{\offcon}{\href{https://github.com/fiorenza2/OffCon3}{\textcolor{pastelred}{OffCon$^3$}}\xspace}

\title{\offcon: What is State-of-the-Art Anyway?}

\author{%
  Philip J. Ball \\
  Department of Engineering Science\\
  University of Oxford\\
  Oxford, UK \\
  \texttt{ball@robots.ox.ac.uk} \\
   \and
   Stephen J. Roberts \\
   Department of Engineering Science \\
   University of Oxford \\
   Oxford, UK \\
   \texttt{sjrob@robots.ox.ac.uk} \\
}

\begin{document}

\maketitle

\begin{abstract}
    % This paper introduces an open-source codebase called \offcon which consolidates two popular model-free off-policy continuous control methods: 1) Soft Actor Critic (SAC); 2) Twin Delayed DDPG (TD3). This allows for easy comparison of these seemingly different methods, both algorithmically and empirically. To demonstrate this, we show that despite many sources showing the relative improvement of SAC over TD3 (often attributed to soft-policy iteration), surprisingly their performance is statistically indistinguishable in most tasks after matching hyperparameters. %As a result, by creating a strong baseline that is closer to traditional off-policy (i.e., non-soft) reinforcement learning, we can more naturally incorporate existing techniques from discrete-control action spaces to improve performance. This results, for the first time to our knowledge, in model-free sample-efficiency in continuous-control domains that is comparable to model-based methods.
    Two popular approaches to model-free continuous control tasks are SAC and TD3. At first glance these approaches seem rather different; SAC aims to solve the entropy-augmented MDP by minimising the KL-divergence between a stochastic proposal policy and a hypotheical energy-basd soft Q-function policy, whereas TD3 is derived from DPG, which uses a deterministic policy to perform policy gradient ascent along the value function. In reality, both approaches are remarkably similar, and belong to a family of approaches we call `Off-Policy Continuous Generalized Policy Iteration'. This illuminates their similar performance in most continuous control benchmarks, and indeed when hyperparameters are matched, their performance can be statistically indistinguishable. To further remove any difference due to implementation, we provide \offcon (\emph{Off}-Policy \emph{Con}tinuous \emph{Con}trol: \emph{Con}solidated), a code base featuring state-of-the-art versions of both algorithms.
\end{abstract}

\section{Introduction}

State-of-the-art performance in model-free continuous control reinforcement learning (RL) has been dominated by off-policy maximum-entropy/soft-policy based methods, namely Soft Actor Critic \cite{haarnoja2018soft, haarnoja2018softapp}. This is evidenced by the plethora of literature, both model-free and model-based, that chooses SAC as the standard \cite{mbpo, clavera2020modelaugmented, chow2020variational, pmlr-v97-rakelly19a, spdg}, often showing it as the best performing model-free approach.

\subsection{Deterministic Policy Gradients}

At this point we introduce the notion of deterministic policy gradients (DPGs) for off-policy reinforcement learning. This is in contrast to stochastic policy gradients (SPGs) that rely on a stochastic policy for gradient estimation. It can be shown that DPG is simply a limiting case of SPG \cite{dpgpaper}, and intuitively the key difference between them focuses on how they each rely on samples for estimating gradients.

For both approaches the policy gradient proof is required. For details see \cite{dpgpaper, suttonbarto}, but through changing the order of integration and/or differentiation we can remove the reliance of the derivative on having access the underlying state distribution.

As aforementioned, DPG is a limiting case of SPG, specifically when the variance parameter of the SPG policy tends to $0$ (i.e., $\sigma \rightarrow 0$). However the similarities and differences between these two methods are nuanced and merit further investigation. This is under-explored and often incorrect equivalences are drawn (i.e., DPG necessitates off-policy learning, SPG necessitates on-policy learning).

% We go into more details in Appendix \ref{app:compare}, but
We start by presenting a simple explanation as to why DPG facilitates off-policy learning:
\begin{align}
    % Q^\pi(s_t, a_t) &= \mathbb{E}_{r_t, s_{t+1} \sim E}\left[ r(s_t,a_t) + \gamma \mathbb{E}_{a_{t+1}\sim\pi}\left[Q^\pi(s_{t+1},a_{t+1}) \right]\right]\label{eq:SPG}\\
    Q(s, a) &= \mathbb{E}_{r, s' \sim E}\left[ r_t + \gamma Q(s',\mu(s'))\right].\label{eq:DPG}
\end{align}
% where Eqs \ref{eq:SPG} and \ref{eq:DPG} loosely represent stochastic and deterministic policy gradient approaches respectively.
Observing Eq \ref{eq:DPG}, we note that the expectation is only dependent on the environment itself, and not the policy. Therefore, all we need to train the Q-function is environment samples (i.e., tuples $(s,a,r,s_{t+1})$ from a replay buffer), and the deterministic policy $\pi$.
We are now in a position to write down the objectives we wish to maximize for both the critic and the actor. For the critic, we use a standard Bellman update, and for the actor, we maximize the expected return under a Q-function:
\begin{align}
    J_Q &= \mathbb{E}_{s,a,r,s'\sim E}\left[ \left(Q(s,a) - \left(r  + \gamma Q(s',a')|_{a'=\pi(s')}\right)\right)^2\right]
\end{align}
For the actor $\pi$, we wish to maximize the expected return:
\begin{align}
    J_\pi &= \mathbb{E}_{s\sim E}\left[V(s)\right]\\
    &= \mathbb{E}_{s\sim E}\left[ Q(s,a)|_{a=\pi(s)}\right]\label{eq:Jactor}
\end{align}
We can now write out the update steps required for DPG-style algorithms, or more specifically DDPG \cite{ddpg} considering the use of neural networks. This will facilitate the comparisons to SAC later on. We now denote the neural network weights of the Q-function and policy as $\theta$ and $\phi$ respectively, with $Q_\theta$ and $\pi(\cdot) = f_\phi(\cdot)$ respectively. We define the policy as a deterministic function for now, as it will make it clearer later when we start defining policy $\pi$ as a distribution that is dependent on a deterministic function. We now write the critic and actor update rules:\\
\textbf{Critic:}
\begin{align}
    \nabla_{\theta} J_Q &\approx \nabla_{\theta} \mathbb{E}_{r,s,s'\sim E}\left[ \left(Q_\theta(s,a) - \left(r  + \gamma Q_\theta(s',a')|_{a'=f_\phi(s')}\right)\right)^2\right]\text{\footnotemark}.
\end{align}
\textbf{Actor:}
\begin{align}
    \nabla_{\phi} J_\pi &\approx \mathbb{E}_{s\sim E}\left[ \nabla_a Q_\theta(s,a)|_{a = f_\phi(s)} \nabla_{\phi} f_\phi(s) \right].
\end{align}
\footnotetext{For completeness, the gradient can be written as $\propto \mathbb{E}_{r,s,s'\sim\rho^\mu}\left[ \nabla_{\theta^Q} Q(s,a|\theta^Q) \left(Q(s,a|\theta^Q) - y_t \right)\right]$ where $y_t$ represents the target network terms} 
Note that, due to the chain rule, the gradient through the actor requires a Q-value estimator that is differentiable with respect to actions.
Finally, we observe that in the generalized policy iteration analogues of Actor-Critic with dynamic programming (\cite{suttonbarto} Chapter 4); critic training is analogous to policy evaluation, and actor training is analogous to policy improvement.

\subsection{Stochastic Value Gradients}

Here we discuss Stochastic Value Gradients \cite{svgpaper}, an algorithm that introduces the idea of taking a gradient through a learned model of the environment and associated stochastic policy. We specifically focus on the model-free limiting case of this approach, SVG(0). This is of particular interest as it represents a stepping stone between DPG and the maximum entropy methods introduced later. We observe that, unlike the other versions of SVG, SVG(0) specifically uses a Q-function to estimate expected return (as opposed to a state-value function). Therefore we must derive the stochastic Bellman equation in the form of the Q-function, following a similar approach to \cite{svgpaper}:
\begin{align}
    Q^\pi(s_t,a_t) &= \mathbb{E}\left[ \gamma^{\tau - t} r^\tau | s=s_t, a=a_t \right] \\
    &= \int r_t p(r_t|s_t,a_t) + \gamma \left[ \int\int Q^\pi(s_{t+1}, a_{t+1}) \pi(a_{t+1}|s_{t+1})p(s_{t+1}|s_t,a_t) \dd{a_{t+1}} \dd{s_{t+1}}\right] \dd{r_t} \\
    &= \mathbb{E}_{r_t, s_{t+1} \sim E}\left[ r_t + \gamma \mathbb{E}_{a_{t+1}\sim\pi}\left[Q^\pi(s_{t+1},a_{t+1}) \right]\right]\label{eq:SPG}.
\end{align}
Observe how Eq \ref{eq:SPG} is just Eq \ref{eq:DPG} except with a stochastic policy $a \sim \pi(\cdot)$. To make its derivative tractable, we treat the policy as a spherical Gaussian, and amortize its parameter ($\mu, \Sigma$) inference using a neural network with weights $\theta$. This allows the use of the reparameterization/pathwise derivative trick \cite{vae, mohamed2020monte}. This means $a \sim \pi(s,\eta;\theta)$ where $\eta \sim \mathcal{N}(0,I)$. As a result, we move policy action sampling outside (i.e., we sample from both the environment $E$ and a $\mathcal{N}(0,I)$), and can backpropagate through the policy weights:
\begin{align}
        Q^\pi(s_t,a_t) &= \mathbb{E}_{r_t, s_{t+1} \sim E}\left[ r_t + \gamma \mathbb{E}_{\eta\sim \mathcal{N}(0,I)}\left[Q^\pi(s_{t+1},\pi(s_{t+1},\eta; \theta)) \right]\right].\\
         &= \mathbb{E}_{r_t, s_{t+1} \sim E, \eta\sim \mathcal{N}(0,I)}\left[ r_t + \gamma\left[Q^\pi(s_{t+1},\pi(s_{t+1},\eta; \theta)) \right]\right]\label{eq:SVG0}.
\end{align}
This means we can write the derivative of the Actor and Critic as follows:\\
\textbf{Critic:}
\begin{align}
    \nabla_{\theta^Q} J &\approx \nabla_{\theta^Q} \mathbb{E}_{r,s,s'\sim\rho^\pi, \eta \sim \mathcal{N}(0,I)}\left[ \left(Q(s,a;\theta^Q) - \left(r  + \gamma Q(s',a';\theta^{Q'})|_{a'=\pi(s',\eta)}\right)\right)^2\right].\\
\end{align}
\textbf{Actor:}
\begin{align}
    \nabla_{\theta^\pi} J &\approx \mathbb{E}_{s\sim\rho^\pi, \eta \sim \mathcal{N}(0,I)}\left[ \nabla_a Q(s,a;\theta^Q)|_{a = \pi(s, \eta)} \nabla_{\theta^\pi} \pi(s,\eta;{\theta^\pi}) \right].
\end{align}
Observe how similar this is to the DPG-style gradients; note that when determining actions an additional sample from a Gaussian distribution is all that is necessary. Furthermore, we observe that this is still an off-policy algorithm, with no dependency on the policy that gave rise to the trajectory samples.

\subsection{Soft Actor Critic}

SAC is an actor-critic method which aims to learn policy that maximizes both return \textbf{and} entropy over each visited state in a trajectory \cite{ziebart}:
\begin{align}
    \pi^* = {\arg \max}_\pi \sum_{t} \mathbb{E}_{(\mathbf{s}_t,\mathbf{a}_t)\sim \rho_\pi} \left[ r(\mathbf{s}_t, \mathbf{a}_t) \begingroup\color{ForestGreen} + \alpha \mathcal{H}(\pi(\cdot|\mathbf{s}_t)) \endgroup \right]\label{eq:softreturn}
\end{align}
where the part of the equation in \textcolor{ForestGreen}{green} describes the additional entropy objective (N.B.: the conventional objective is therefore recovered as $\alpha \rightarrow 0$). This is done using soft-policy iteration, and involves repeatedly applying the following entropy Bellman operator \cite{haarnoja2018soft}:
\begin{align}
    \mathcal{T}^\pi Q(s,a) =& r(s,a) + \gamma \mathbb{E}_{s'\sim p} [ V(s') ]
\end{align}
where
\begin{align}
    V(s) =& \mathbb{E}_{a \sim \pi} [ Q(s,a) - \alpha \log\pi(a|s)].\label{eq:softV}
\end{align}
For consistency of presentation, we present this as a (soft) Bellman update:
\begin{align}
    Q^\pi(s,a) &= \int r p(r|s,a) + \gamma \left[ \iint (Q^\pi(s', a') - \alpha \log \pi(a'|s')) \pi(a'|s')p(s'|s,a) \dd{a'} \dd{s'}\right] \dd{r} \nonumber \\
    &= \mathbb{E}_{r, s' \sim E}\left[ r + \gamma \mathbb{E}_{a'\sim\pi}\left[Q^\pi(s',a')-\alpha \log \pi(a'|s') \right]\right] \nonumber \\
     &= \mathbb{E}_{r, s' \sim E, \eta\sim \mathcal{N}(0,I)}\left[ r + \gamma\left[Q^\pi(s',\pi(s',\eta; \theta)) -\alpha \log \pi(s',\eta; \theta) \right]\right]
\end{align}
where in the last line we make the same assumption about amortizing the policy distribution as in SVG.

At this point we can directly write down the objective of the actor/policy, namely to maximize expected return \emph{and} entropy, i.e., Eq \ref{eq:softV}. This follows the method for determining the objective function for the policy gradient in DPG (Eq \ref{eq:Jactor}):
\begin{align}
    J_\pi &= \mathbb{E}_{s\sim\rho^\mu}\left[V(s)\right]\\
    &= \mathbb{E}_{s\sim\rho^\mu}\left[ \mathbb{E}_{a\sim\pi}\left[Q(s,a;\theta^Q) - \alpha \log\pi(a|s)\right]\right]\label{eq:JSactor}
\end{align}
Similarly for the critic $Q$, we have $J_Q$:
\begin{align}
    J_Q &= \mathbb{E}_{r,s,s'\sim\rho^\mu, a'\sim\pi}\left[ \left(Q(s,a;\theta^Q) - \left(r  + \gamma(Q(s',a';\theta^{Q'}) - \alpha \log \pi(a'|s'))\right)\right)^2\right]
\end{align}
The `soft' critic gradient is similar to the DPG style update as the Q-value parameters don't depend on the additional entropy term, however the actor gradient requires both the chain rule and the law of total derivatives. 
%We show details of the calculation in Appendix \ref{app:deriveSACPolicyGrad}.
Here we write down the gradients directly:\\
\textbf{Critic:}
\begin{align}
    \nabla_{\theta^Q} J &\approx \nabla_{\theta^Q} \mathbb{E}_{r,s,s'\sim\rho^\pi, \eta \sim \mathcal{N}(0,I)}\left[ \left(Q(s,a;\theta^Q) - \left(r  + \gamma (Q(s',a';\theta^{Q'}) - \alpha \log \pi(a'|s'))|_{a'=\pi(s',\eta)}\right)\right)^2\right].
\end{align}
\textbf{Actor:}
\begin{align}
    \nabla_{\theta^\pi} J &\approx \mathbb{E}_{s\sim\rho^\pi, \eta \sim \mathcal{N}(0,I)}\left[ \left(-\nabla_{\theta^\pi} \log \pi(a|s) + \nabla_a \left( Q(s,a;\theta^Q) - \alpha \log \pi(s,\eta;\theta^\pi) \right)\right)|_{a = \pi(s, \eta)} \nabla_{\theta^\pi} \pi(s,\eta;{\theta^\pi}) \right].
\end{align}
What remains to be optimized is the temperature $\alpha$, which balances the entropy/reward trade-off in Eq \ref{eq:softreturn}. In \cite{haarnoja2018softapp} the authors learn this during training using an approximation to constrained optimization, where the mean trajectory entropy $\mathcal{H}$ is the constraint.

\subsection{DPG \texorpdfstring{$\rightarrow$}{} SVG \texorpdfstring{$\rightarrow$}{} SAC}

Having outlined DPG, SVG(0), and SAC we are now in a position to directly compare all three approaches. We do this by observing the Critic and Actor objectives, highlighting in different colors the components that are attributable to each:
\begin{align}
    J_\pi &= \mathbb{E}_{s\sim E, \begingroup\color{magenta}\eta\sim\mathcal{N}(0,I)\endgroup}\left[ Q_\theta(s,a) \begingroup\color{ForestGreen}- \alpha \log\pi(a|s)\endgroup|_{a= f_\phi(s,\begingroup\color{magenta}\eta\endgroup)}\right]\\
    J_Q &= \mathbb{E}_{r,s,s'\sim E, \begingroup\color{magenta}\eta\sim\mathcal{N}(0,I)\endgroup}\left[ \left(Q_\theta(s,a) - \left(r  + \gamma(Q_\theta(s',a') \begingroup\color{ForestGreen}- \alpha \log \pi(a'|s'))\endgroup\right)\right)^2|_{a'= f_\phi(s',\begingroup\color{magenta}\eta\endgroup)}\right]
\end{align}
where terms in \textcolor{magenta}{pink} are introduced by SVG(0), and terms in \textcolor{ForestGreen}{green} are introduced by SAC. Here we describe the natural progression of DPG to SAC:
\begin{enumerate}
    \item \textbf{DPG} introduces the policy iteration framework, including the deterministic policy gradient, that allows the learning of policies through Q-learning over continuous action spaces. \textbf{DDPG} introduces heuristics that allow the use of neural network function approximators.
    \item \textbf{SVG} introduces the idea of stochastic policies, and its limiting model-free case \textbf{SVG(0)} allows the learning of stochastic policies in the Q-learning policy improvement framework proposed in DPG. This uses the pathwise derivative through the amortized Gaussian policy.
    \item \textbf{SAC} leverages the policy variance learning in amortized inference by ensuring a maximum-entropy action distribution for any given state through the addition of an entropy term into the traditional maximum return objective.
\end{enumerate}
We observe therefore that all three algorithms can be considered as belonging to the same family, namely `Off-Policy Continuous Generalized Policy Iteration', where the policy evaluation step represents a gradient step along $J_Q$, and policy improvement a gradient step along $J_\pi$. All that distinguishes these approaches is whether the actor is deterministic, and whether there is an additional entropy objective. We note that the SAC policy has been derived using standard gradient ascent of the value function (as in \cite{dpgpaper}), and similarly the DPG policy gradient can be derived as a KL-minimization (as in \cite{haarnoja2018soft}).

\section{Practical Reinforcement Learning}

Two methods derived from the aforementioned approaches have emerged as being most popular, namely SAC with entropy adjustment \cite{haarnoja2018softapp} and the DDPG derived TD3 \cite{td3}. At first glance, it may appear coincidental that both approaches have achieved similar levels of success in continuous control tasks, such as OpenAI Gym \cite{opeanigym}, but the above analysis shows that they are closely related. We briefly explain the merits of TD3, and understand how this has influenced SAC.

\subsection{TD3}
TD3 \cite{td3} is based on DDPG, and introduces several heuristics to improve upon it. These include:
\begin{itemize}
    \item Training two Q-functions, then taking their minimum when evaluating to address Q-function overestimation bias.
    \item Update target parameters and actor parameters less frequently than critic updates.
    \item Add noise to the target policy action during critic training, making it harder for the actor to directly exploit the critic.
\end{itemize}
The original SAC paper \cite{haarnoja2018soft} does not train two Q-functions, and instead trains a Q-function and a state-value (V) function. Furthermore the trade-off between entropy and reward is fixed. The `applied' SAC paper \cite{haarnoja2018softapp} removes the state-value function, and instead trains two Q-functions similar to TD3, and automatically adjusts the temperature trade-off to ensure some expected policy entropy (a function of action dimension). Interestingly, in their original papers, TD3 and SAC claim to outperform each other, and it would appear that the incorporation of the TD3-style Q-learning and temperature adjustment results in the ultimately better performance in the `applied' SAC paper. However there are still key differences between SAC and TD3 training, namely heuristics such as network architecture, learning rate, and batch size. For the purposes of fair comparison, we choose these to be the same across both SAC and TD3, as shown in Table \ref{tab:hyperparams}\footnote{Note we include an additional hidden layer, see Appendix \ref{app:shallow} for details}.

\begin{table}[ht]
    \centering
\begin{tabular}{lcc}\toprule
& \multicolumn{2}{c}{Algorithm}
\\\cmidrule(lr){2-3}
 Hyperparamater          & TD3 & SAC \\\midrule
Collection Steps & \multicolumn{2}{c}{1,000}  \\
Random Action Steps& \multicolumn{2}{c}{10,000} \\
Network Hidden Layers & \multicolumn{2}{c}{$256:256:256$} \\
Learning Rate&  \multicolumn{2}{c}{\num{3e-4}}  \\
Optimizer &\multicolumn{2}{c}{$Adam$} \\
Replay Buffer Size & \multicolumn{2}{c}{\num{1e6}} \\
Action Limit & \multicolumn{2}{c}{$[-1,1]$} \\
Exponential Moving Averaging Parameter & \multicolumn{2}{c}{\num{5e-3}} \\
(Critic Update:Environment Step) Ratio & \multicolumn{2}{c}{1} \\
(Policy Update:Environment Step) Ratio & 2 & 1 \\
Has Target Policy? & Yes & No \\
Expected Entropy Target & N/A & $-\text{dim}(\mathcal{A})$ \\
Policy Log-Variance Limits & N/A & [-20, 2] \\
Target Policy $\sigma$ & 0.2 & N/A \\
Target Policy Clip Range & [-0.5, 0.5] & N/A \\
Rollout Policy $\sigma$ & 0.1 & N/A\\
\bottomrule
\end{tabular}
\caption{Hyperparameters used in \offcon}
\label{tab:hyperparams}
\end{table}

\subsection{What is the effect of Gaussian exploration?}
\label{subsec:gaussian}
One difference between TD3 and DDPG is the noise injection applied during Q-value function training. This turns a previously deterministic mapping $a = \mu(s)$ into a stochastic one ($a \sim \mu(s) + \text{clip}(\mathcal{N}(0,I)\times0.2,-0.5,0.5)$). This means that the policies used in both data collection and critic training are in fact stochastic, making TD3 closer to SAC. We may ask how this compares to a deterministic objective; evidently, veering from the mean action selected by the deterministic actor should reduce expected return, so what does this stochasticity provide? To explore this question, we split our analysis into two sections: the effect on the Critic, and the effect on the Actor.
\paragraph{Effect on Critic:}
We simplify analysis by assuming all added noise is diagonal Gaussian\footnote{Action clipping technically makes this assumption untrue, but in reality policies are still very nearly Gaussian}, and write the deterministic objective as $J_D$. We also assume 1-D actions without loss of generality. Performing a Taylor series expansion of the stochastic policy, we find that the objective maximized by this Gaussian actor ($J_R$) as (see Appendix \ref{app:objective} for proof):
\begin{align}
J_R &\approx J_D + \frac{\sigma^2}{2} \mathbb{E}_{s_t\sim E} \left[ \nabla^2_a Q(s_t,a)|_{a = \mu(s_t)} \right]
\end{align}
This is the deterministic objective with an additional term proportional to the fixed variance of the policy, as well as the 2nd derivative (Hessian for multi-dimensional actions) of the critic with respect to actions. Unpacking this latter term, noting that the following residual between the stochastic ($J_R$) and deterministic ($J_D$) objectives, leads to:
\begin{align}
J_R - J_D \approx \frac{\sigma^2}{2} \mathbb{E}_{s_t\sim E} \left[ \nabla^2_a Q(s_t,a)|_{a = \mu(s_t)} \right]
\end{align}
First, let us consider a well trained policy that is able to produce actions that maximize the critic $Q$ for all states. This means that the value of the 2nd order term must be negative (equivalently, the Hessian must be negative semi-definite). Evidently, any non-zero $\sigma^2$ will result in the stochastic return $J_D$ being lower than $J_R$. This implies that the stochastic policy objective $J_R$ can only ever realistically lower bound the deterministic objective $J_D$.

However in Gaussian exploration we fix this $\sigma^2$ to be non-zero, therefore the only degree of freedom is in the second-order term itself. Evidently we want a policy that maximizes $Q$ (i.e., 0th order term), therefore making this term positive is not viable. However the magnitude of the second-order term can be reduced by making $Q$ `smoother'. Since $Q$ is twice differentiable w.r.t. $a$, we can invoke the identity $\nabla^2_a Q(s,a) \preceq \beta \text{I}$ \cite{bubeckoptimize}, implying that the largest eigenvalue of the Hessian of $Q$ is smaller than $\beta$, where $\beta$ is the Lipschitz constant of $Q$. Therefore, to minimize the magnitude of $\nabla^2_a Q(s,a)$, we must learn a $Q$ that is smoother with respect to actions. This can be viewed as a spectral norm regularizer of the $Q$ function \cite{yoshida2017spectral}, and the mechanism that is used in \cite{td3} to ensure the stability of the critic can be viewed as approximating this approach. It must be noted that we get this smoothing behavior by default in SAC as the learned policy has non-zero variance due to its entropy objective.

\paragraph{Effect on Actor:}
The forced non-zero $\sigma^2$ variance term also has implications to entropy based approaches. SAC learns a non-zero variance to ensure some minimum entropy per trajectory timestep:
\begin{align}
    \max_{\pi} \left[ Q^\pi(s_0, a_0) \right] \quad \text{s.t.} \quad \mathbb{E}_{s_t\sim E} [ \mathbb{E}_{a_t \sim \pi} [ - \log \pi(a_t|s_t) ]] > \mathcal{H} \quad \forall t
\end{align}
where, for a Gaussian policy, we can write
\begin{align}
    \max_{\pi} \left[ Q^\pi(s_0, a_0) \right] \quad \text{s.t.} \quad \mathbb{E}_{s_t\sim E} \left[ \log (\sigma(s_t) \sqrt{2\pi e}) \right] > \mathcal{H} \quad \forall t.
\end{align}
This optimization is non-trivial for SAC (see Sec.\ 5 in \cite{haarnoja2018softapp}) as the amount of policy variance is learned by the policy (hence the $s_t$ dependency above). However for a policy with fixed variance $\sigma$, the optimization becomes trivial as we can drop the state dependency, and simply maximize policy over the critic\footnote{Consider that for a fixed $\sigma$ we can always find an $\mathcal{H}$ such that the constraint inequality evaluates to an exact equality for all policies, therefore the dual simply collapses into a maximization of the primal without a constraint.} (which is done in standard actor training):
\begin{align}
    \max_{\pi} \left[ Q^\pi(s_0, a_0) \right] \quad \text{s.t.} \quad  \log (\sigma \sqrt{2\pi e}) > \mathcal{H} \quad \forall t.
\end{align}
In the case of a deployed policy which has standard deviation $0.1$, such as in TD3, we can view this performing exploration with a maximum entropy policy that ensures a policy entropy of $\approx -0.884$ nats.

\subsection{Why not SVG(0)?}
\begin{figure}[H]
\centering
    \includegraphics[width=0.45\textwidth]{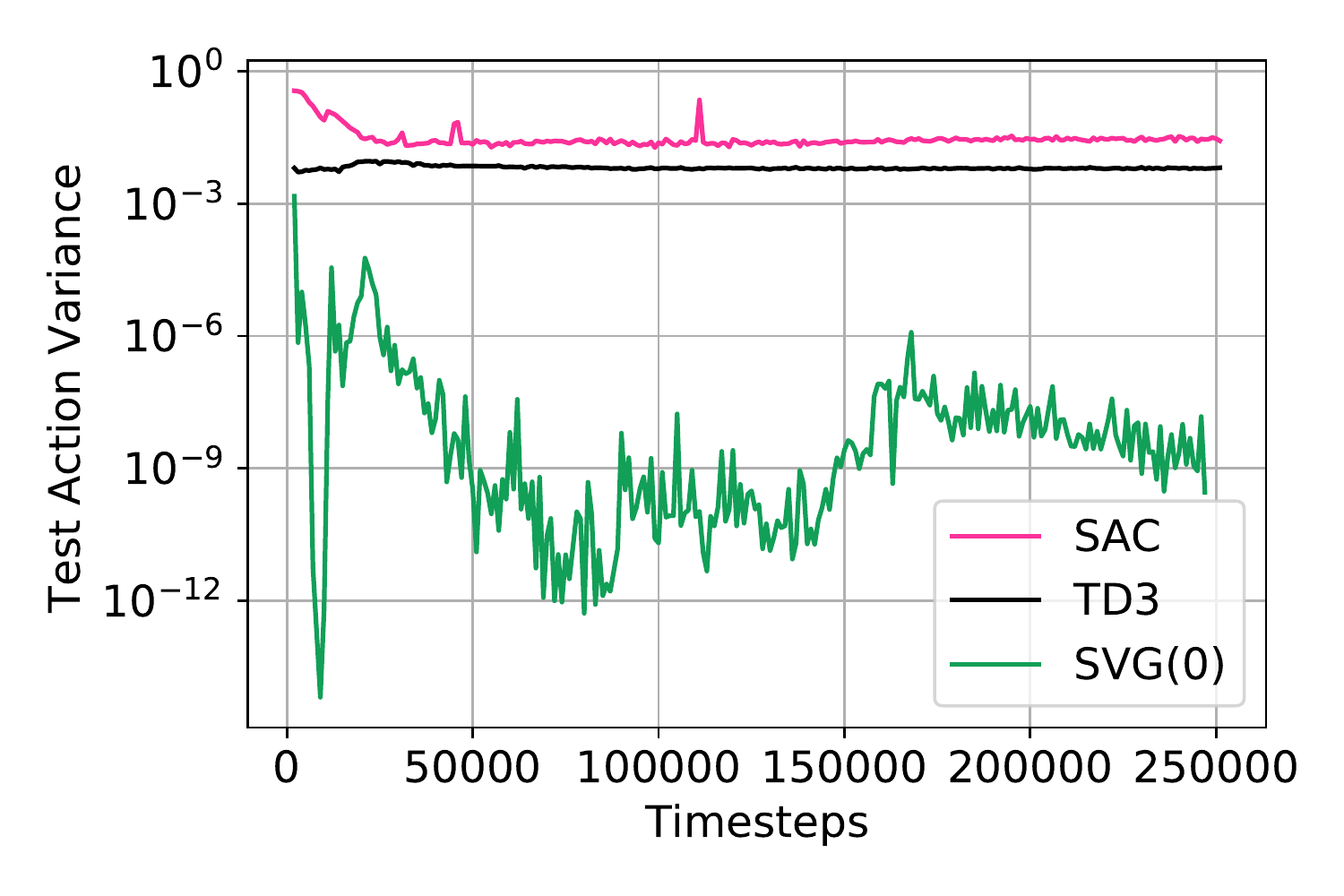}
\caption{Variance of rollout policies on HalfCheetah}
\label{fig:var}
\end{figure}

In principle, SVG(0) appears to be a strong candidate for policy training; it is stochastic, does not incorporate an entropy term, which adds computation and hyperparameters. In reality, since the environments evaluated are deterministic, the variance head of the SVG(0) policy tends to 0 very quickly. The reason for this is outlined in Sec.\ \ref{subsec:gaussian}. As a consequence, the resultant algorithm is effectively DDPG. Indeed this is supported in the analysis performed in Appendix \ref{app:objective}, where only the 0th order Thompson term remains for relatively small $\sigma$. We illustrate this effect in Figure \ref{fig:var}, and include this algorithm for illustrative purposes as `TDS' in \offcon.

\section{Experiments}

For these experiments, we run both algorithms for 5 seeds on 4 different MuJoCo environments: HalfCheetah, Hopper, Walker2d, and Ant. We then perform a two-tailed Welch's $t$-test \cite{welch1947} to determine whether final performance is statically significantly different. Observing Table \ref{tab:ttest}, Ant and Walker2d performance is statistically indistinguishable (although TD3 learns quicker in Ant, see Appendix \ref{app:efficiencygain}); in HalfCheetah SAC does convincingly outperform TD3, but in Hopper, the opposite is true.

\begin{figure}[H]
\centering
\begin{subfigure}{.45\textwidth}
    \centering
    \includegraphics[width=\textwidth]{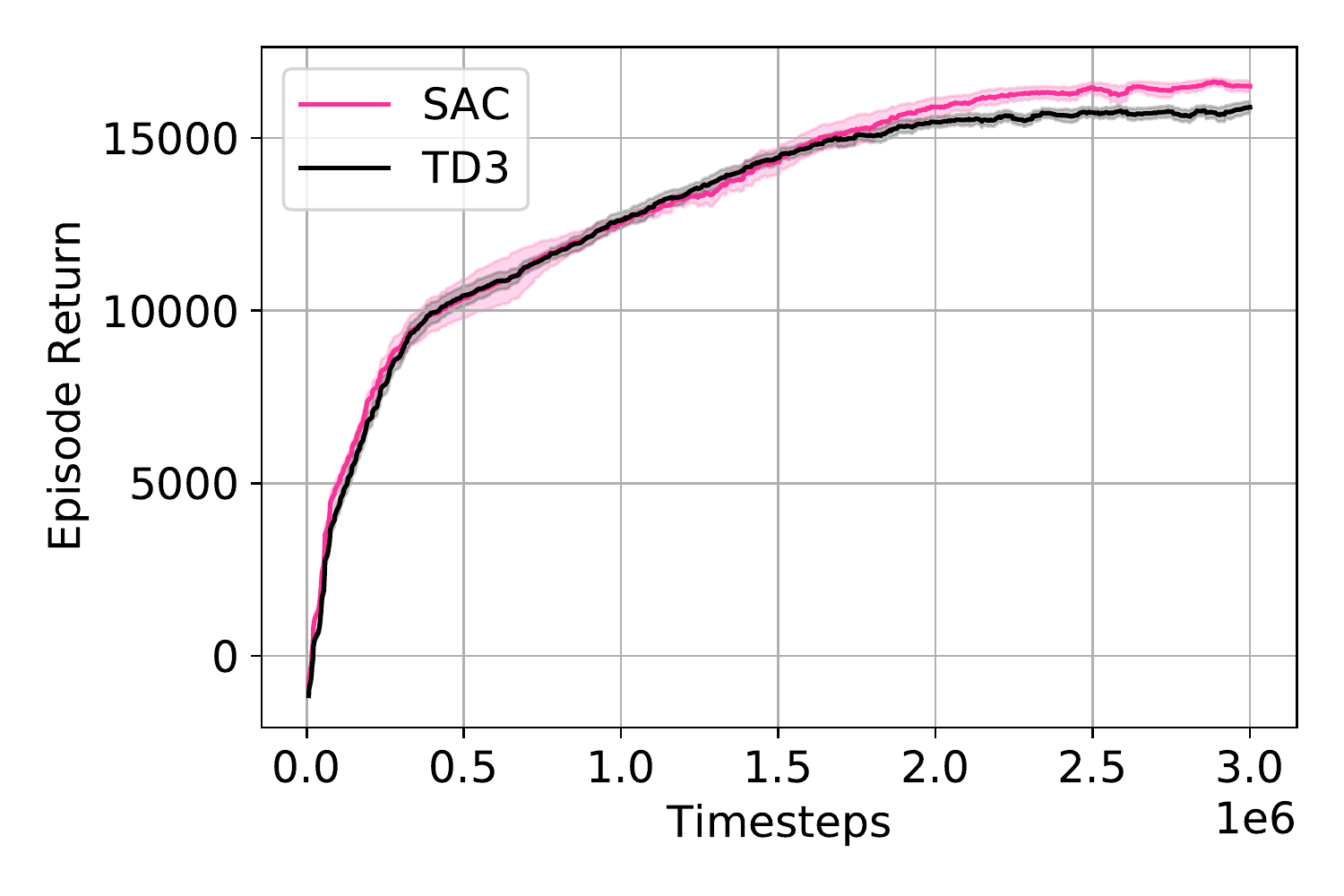}
    \caption{HalfCheetah}
    \label{fig:halfcheetah}
\end{subfigure}
\hspace{0.5cm}
\begin{subfigure}{.45\textwidth}
    \centering
    \includegraphics[width=\textwidth]{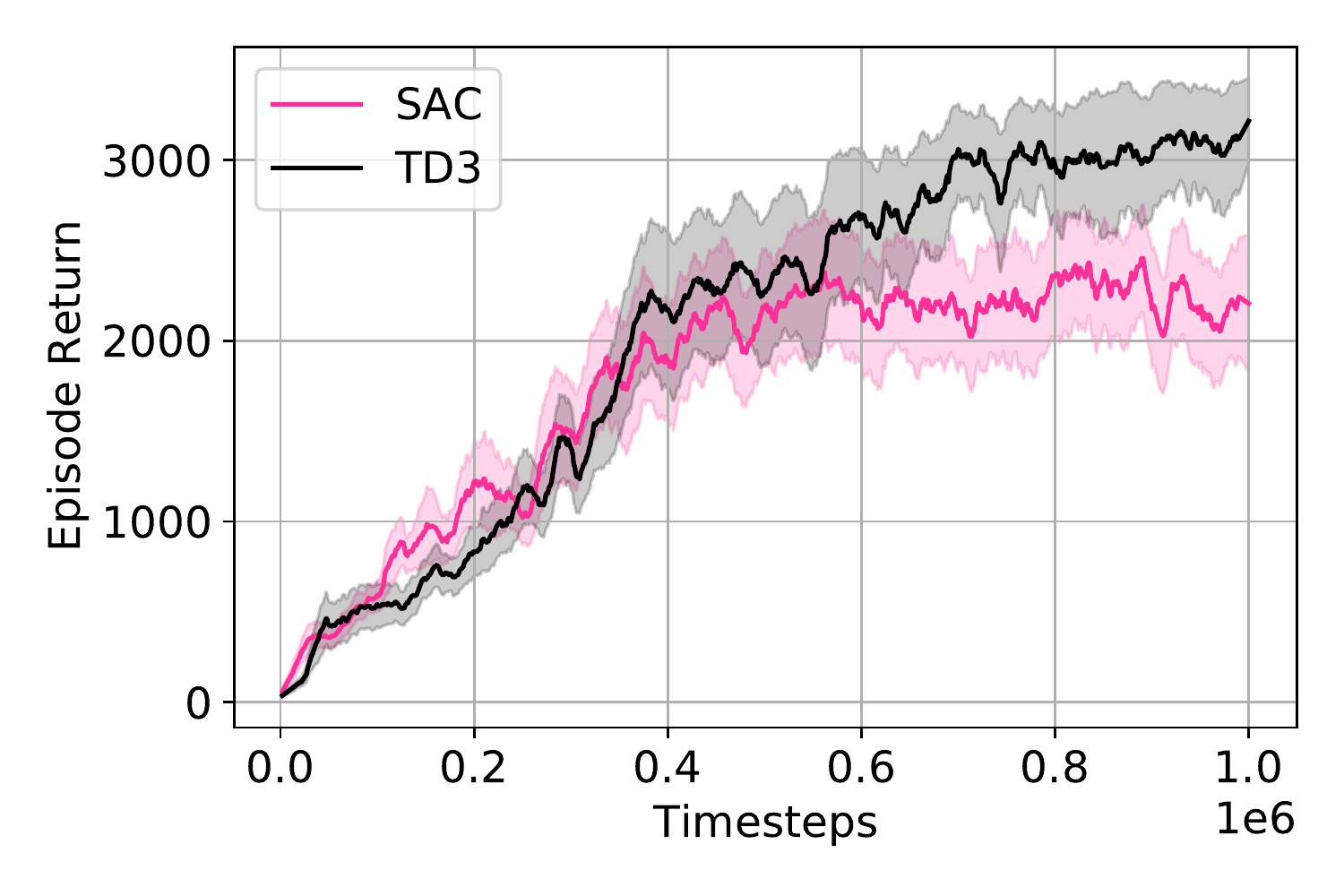}
    \caption{Hopper}
    \label{fig:hopper}
\end{subfigure}
\begin{subfigure}{.45\textwidth}
    \centering
    \includegraphics[width=\textwidth]{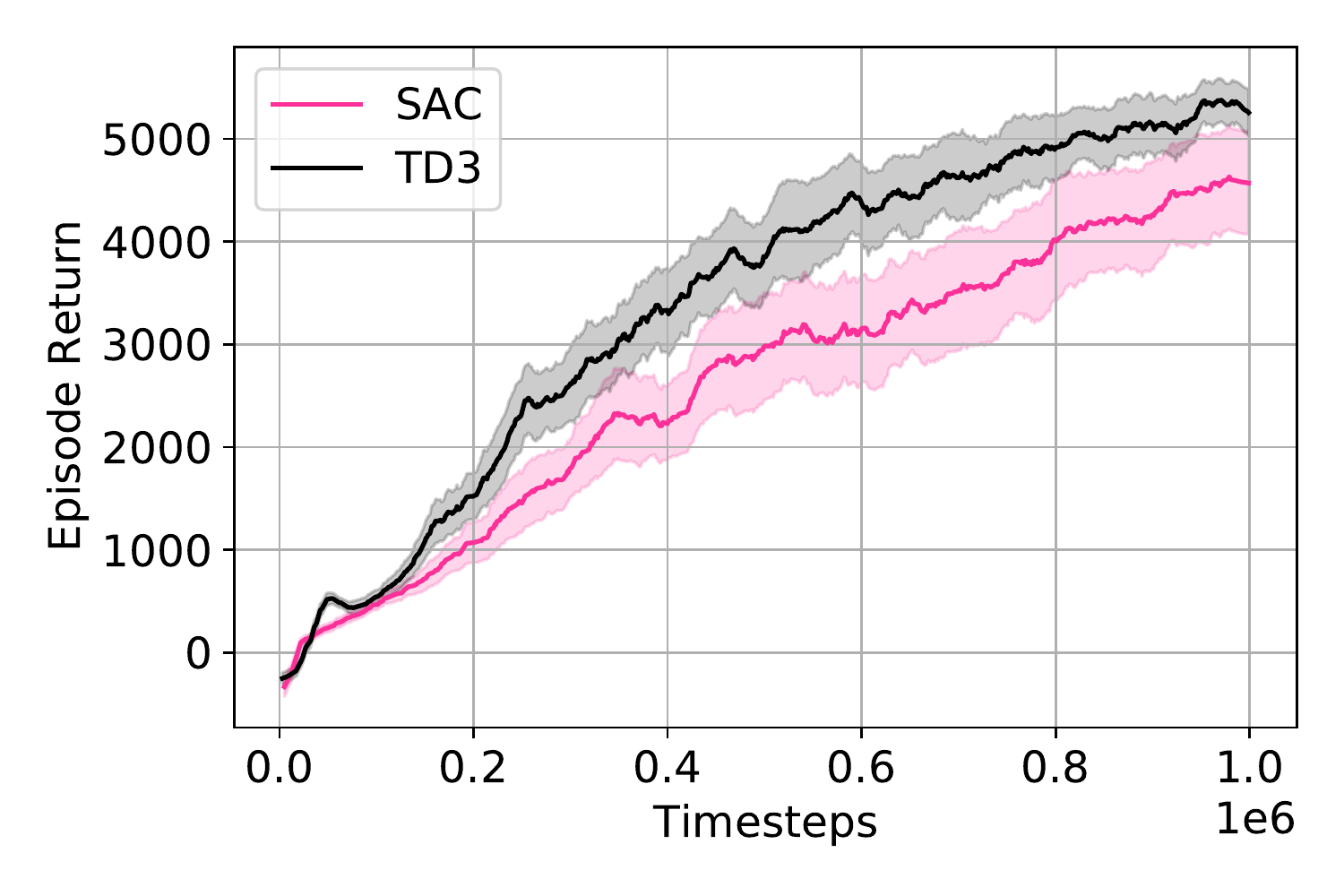}
    \caption{Ant}
    \label{fig:ant}
\end{subfigure}
\begin{subfigure}{.45\textwidth}
    \centering
    \includegraphics[width=\textwidth]{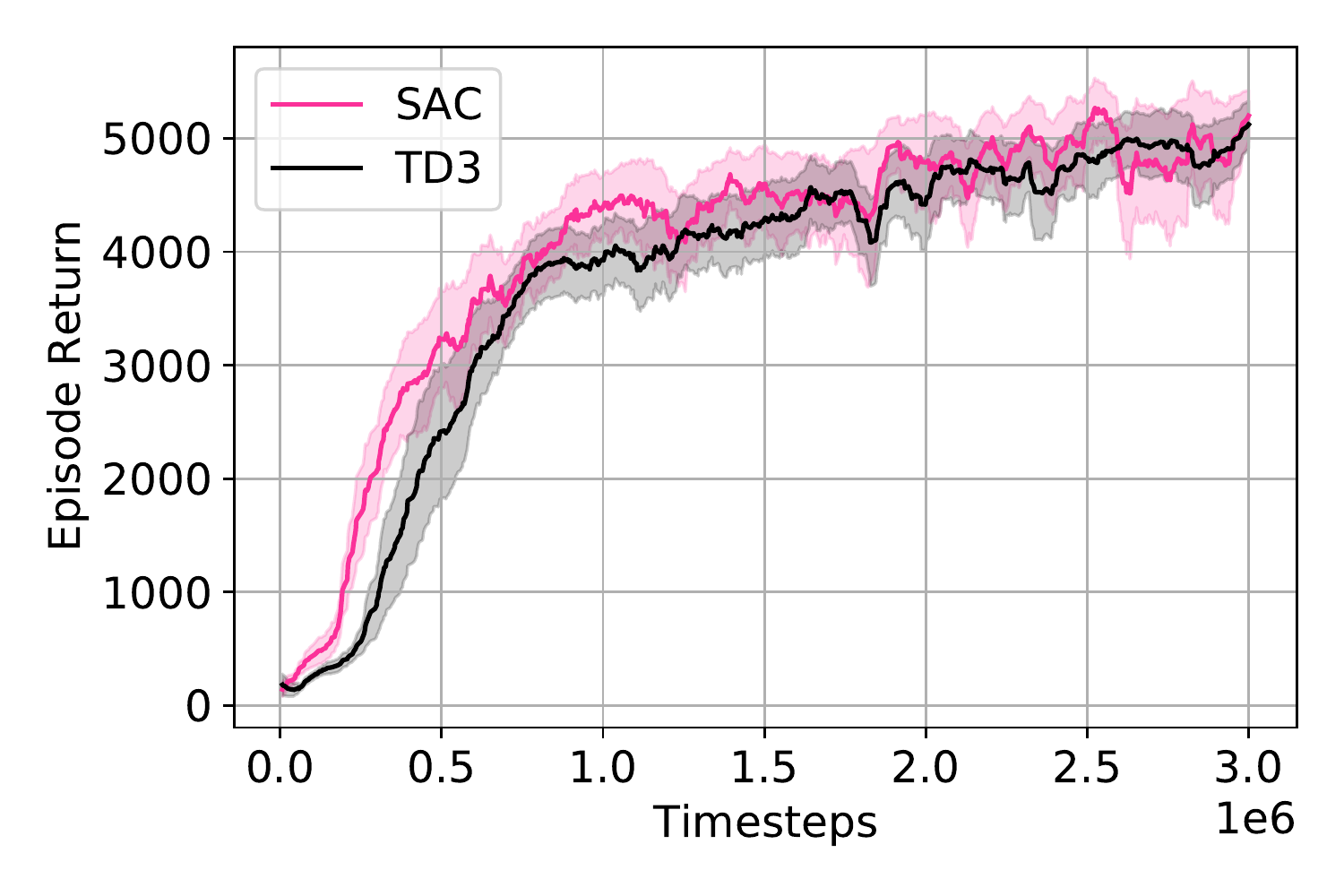}
    \caption{Walker2d}
    \label{fig:walker}
\end{subfigure}
\begin{subfigure}{.45\textwidth}
    \centering
    \includegraphics[width=\textwidth]{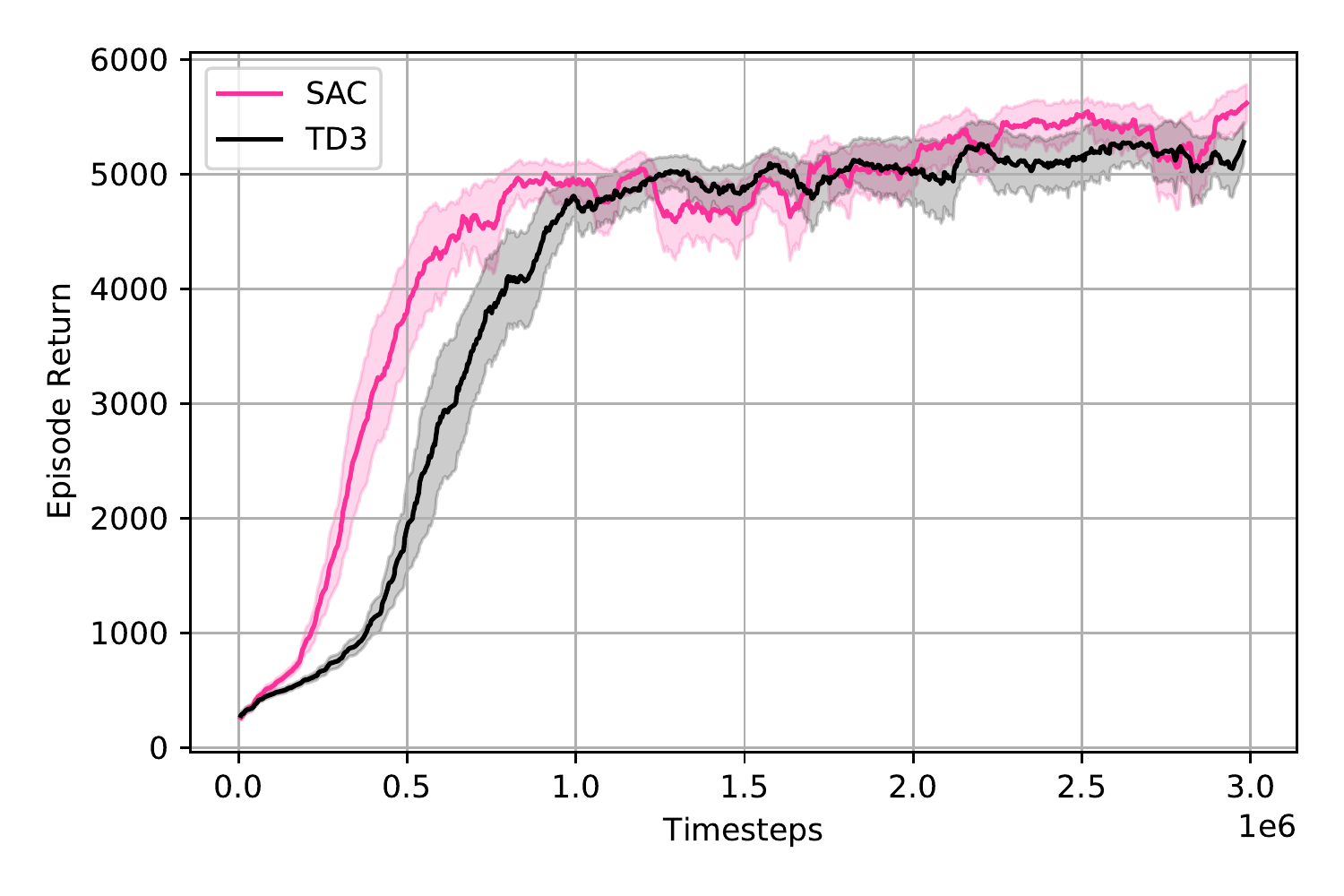}
    \caption{Humanoid}
    \label{fig:walker}
\end{subfigure}
\caption{SAC and TD3 Training Curves on MuJoCo Environments.}
\label{fig:compare}
\end{figure}%

\subsection{Authors' Results}
For completeness we compare the results from each algorithm's available code with ours to ensure our implementation does not unfairly penalize either approach. The results are shown in Tables \ref{tab:compareOGSAC} and \ref{tab:compareOGTD3}\footnote{Authors' SAC and TD3 code is \href{https://github.com/haarnoja/sac}{here} and \href{https://github.com/sfujim/TD3}{here} respectively. We tabulate the results provided in these repos. Results are the max (averaged over seeds) performance up to and including the Timesteps column.}. We note that our implementation appears to generally match, or exceeds, the authors' code. Note that unlike the original code in \cite{td3}, we do not discard `failure' seeds\footnote{See discussion \href{https://openreview.net/forum?id=SJgn464tPB}{here}.}; this may explain why our implementation doesn't always outperform the author's code, especially on less stable environments (such as Hopper and Walker2d).

\begin{table}[h]
    \centering
\begin{tabular}{lcc}\toprule
& \multicolumn{2}{c}{$t$-Test Result}
\\\cmidrule(lr){2-3}
 Environment          & $t$ & $p$ \\\midrule
HalfCheetah    & 4.29 & \begingroup\color{OrangeRed}0.00927\endgroup \\
Hopper& -2.92 & \begingroup\color{OrangeRed}0.0293\endgroup\\
Ant & -0.481 & 0.653\\
Walker2d& 1.59 & 0.155 \\
Humanoid& 1.29 & 0.29 \\
\bottomrule
\end{tabular}
\caption{Two-tailed Welch's $t$-test results}
\label{tab:ttest}
\end{table}

\begin{table}[H]
    \centering
\begin{tabular}{lccc}\toprule
& \multicolumn{2}{c}{SAC Return} &
\\\cmidrule(lr){2-3}
 Environment          & Ours & Author & Timesteps \\\midrule
HalfCheetah    & $16,784\pm292$ & $12,219\pm4,899$ &\num{3e6}\\
Hopper& $3,142\pm654$ & $3,319\pm175$ &\num{1e6}\\
Ant & $4,987\pm784$ & $3,845\pm759$&\num{1e6}\\
Walker2d& $5,703\pm408$ & $5,523\pm466$& \num{3e6}\\
Humanoid& $5,871\pm171$ & $6,268\pm186$& \num{3e6}\\
\bottomrule
\end{tabular}
\caption{SAC Implementation Comparison to Author's Code}
\label{tab:compareOGSAC}
\end{table}

\begin{table}[H]
    \centering
\begin{tabular}{lccc}\toprule
& \multicolumn{2}{c}{TD3 Return} &
\\\cmidrule(lr){2-3}
 Environment          & Ours & Author & Timesteps \\\midrule
HalfCheetah    & $12,804\pm493$ & $9,637\pm859$ & \num{1e6}\\
Hopper& $3,498 \pm 99$ & $3,564\pm115$ &\num{1e6}\\
Ant & $5,700\pm334$ & $4,372\pm1,000$&\num{1e6}\\
Walker2d& $4,181\pm607$ & $4,683\pm540$& \num{1e6}\\
Humanoid& $5,085\pm144$ & N/A & \num{1e6}\\
\bottomrule
\end{tabular}
\caption{TD3 Implementation Comparison to Author's Code}
\label{tab:compareOGTD3}
\end{table}

\section{Conclusion}

In conclusion, we show that TD3 and SAC are closely related algorithms, and that it is possible to categorize them as belonging to the same general family of algorithms, namely `Off-Policy Continuous Generalized Policy Iteration’. We make this comparison complete by comparing against an oft-forgotten approach SVG(0). We then show that by matching hyperparameters, their performance is more similar than is often shown in the literature, and can be statistically indistinguishable; furthermore TD3 can in fact outperform SAC on certain environments whilst being more computationally efficient. To make this link from theory to practice explicit, we have implemented both in the open-source code base \offcon, whereby many major elements of the code are shared for each algorithm.

\printbibliography

\begin{appendix}
% \section{Q-function Actor-Critic v.s. V-function Actor-Critic}
% \label{app:compare}

% \section{Deriving the Soft Policy Gradient}
% \label{app:deriveSACPolicyGrad}

\section{Objective of a Stochastic Gaussian Policy with fixed Variance}
\label{app:objective}

Consider a Deterministic Policy:
\begin{align*}
J_{D} &= \mathbb{E}_{s_t\sim E}\left[ Q(s_t,a_t)|_{a_t= \mu_{(s_t)}}\right].
\end{align*}
The Random Policy is defined as $\pi(a_t|s_t) = \mathcal{N}(a_t|\mu(s_t),\sigma^2)$ where $\sigma$ is fixed:
\begin{align*}
J_{R} &= \mathbb{E}_{s_t\sim E}\left[ \mathbb{E}_{a_t \sim \pi(a_t|s_t)}\left[ Q(s_t,a_t)\right]\right]\\
&= \mathbb{E}_{s\sim E} \left[ \int \mathcal{N}(a_t|\mu(s_t),\sigma^2) Q(s_t,a_t) da_t\right].
\end{align*}
Performing a Taylor expansion of $Q(s_t, a_t)$ around $a_t = \mu(s)$ provides:
\begin{multline*}
    Q(s_t,a_t) = Q(s_t,\mu(s_t)) + \nabla_a Q(s_t,a)|_{a = \mu(s_t)}(a_t - \mu(s_t)) \\ + \frac{1}{2}\nabla^2_a Q(s_t,a)|_{a=\mu(s_t)}(a_t-\mu(s_t))^2 + \dots.
\end{multline*}
We address the different Taylor expansion orders separately (labeled \Circled{0}, \Circled{1}, \Circled{2}, etc.).\\
First 0th order:
\begin{align*}
\text{\Circled{0}} &= \mathbb{E}_{s_t\sim E} \left[ \int \mathcal{N}(a_t|\mu(s),\sigma^2) Q(s_t,\mu(s_t)) da_t \right] \\
&= \mathbb{E}_{s_t\sim E} \left[ Q(s_t,\mu(s_t)) \int \mathcal{N}(a_t|\mu(s_t),\sigma^2) da_t \right] \\
&= \mathbb{E}_{s_t\sim E} \left[ Q(s_t,\mu(s_t)) \right]\\
&= J_D.
\end{align*}
Now 1st order:
\begin{align*}
\text{\Circled{1}} &= \mathbb{E}_{s_t\sim E} \left[ \int \mathcal{N}(a_t|\mu(s_t),\sigma^2) \left( \nabla_a Q(s_t,a)|_{a = \mu(s_t)}(a_t - \mu(s))\right) da_t\right]\\
&= \mathbb{E}_{s_t\sim E} \left[ \nabla_a Q(s_t,a)|_{a = \mu(s_t)} \int \mathcal{N}(a_t|\mu(s_t),\sigma^2) \left( a_t - \mu(s_t)\right) da_t\right]\\
&= \mathbb{E}_{s_t\sim E} \left[ \nabla_a Q(s_t,a)|_{a = \mu(s_t)} \left( \int \mathcal{N}(a_t|\mu(s_t),\sigma^2) a_t da_t - \mu(s_t) \right) \right] \\
&= \mathbb{E}_{s_t\sim E} \left[ \nabla_a Q(s_t,a)|_{a = \mu(s_t)} \left( \mu(s_t) - \mu(s_t) \right) \right]\\
&= 0.
\end{align*}
Now 2nd order:
\begin{align*}
\text{\Circled{2}} &= \mathbb{E}_{s_t\sim E} \left[ \int \mathcal{N}(a_t|\mu(s_t),\sigma^2) \left( \nabla^2_a Q(s_t,a)|_{a = \mu(s_t)}(a_t - \mu(s))^2\right) da_t\right]\\
&= \mathbb{E}_{s_t\sim E} \left[ \nabla^2_a Q(s_t,a)|_{a = \mu(s_t)} \int \mathcal{N}(a_t|\mu(s_t),\sigma^2) \left( a_t - \mu(s))^2\right) da_t\right]\\
&= \mathbb{E}_{s_t\sim E} \left[ \nabla^2_a Q(s_t,a)|_{a = \mu(s_t)} \mathbb{E}_{a_t}\left[ (a_t - \mu(s))^2 \right]\right]\\
&= \mathbb{E}_{s_t\sim E} \left[ \nabla^2_a Q(s_t,a)|_{a = \mu(s_t)} \right] \sigma^2
\end{align*}
So putting it all (\Circled{0}, \Circled{1}, \Circled{2}) together:
\begin{align*}
J_R &= J_D + \frac{\sigma^2}{2} \mathbb{E}_{s_t\sim E} \left[ \nabla^2_a Q(s_t,a)|_{a = \mu(s_t)} \right]
\end{align*}

\section{Efficiency Gain of TD3 in Ant}
\label{app:efficiencygain}
\begin{figure}[h]
    \centering
    \includegraphics[width=0.5\textwidth]{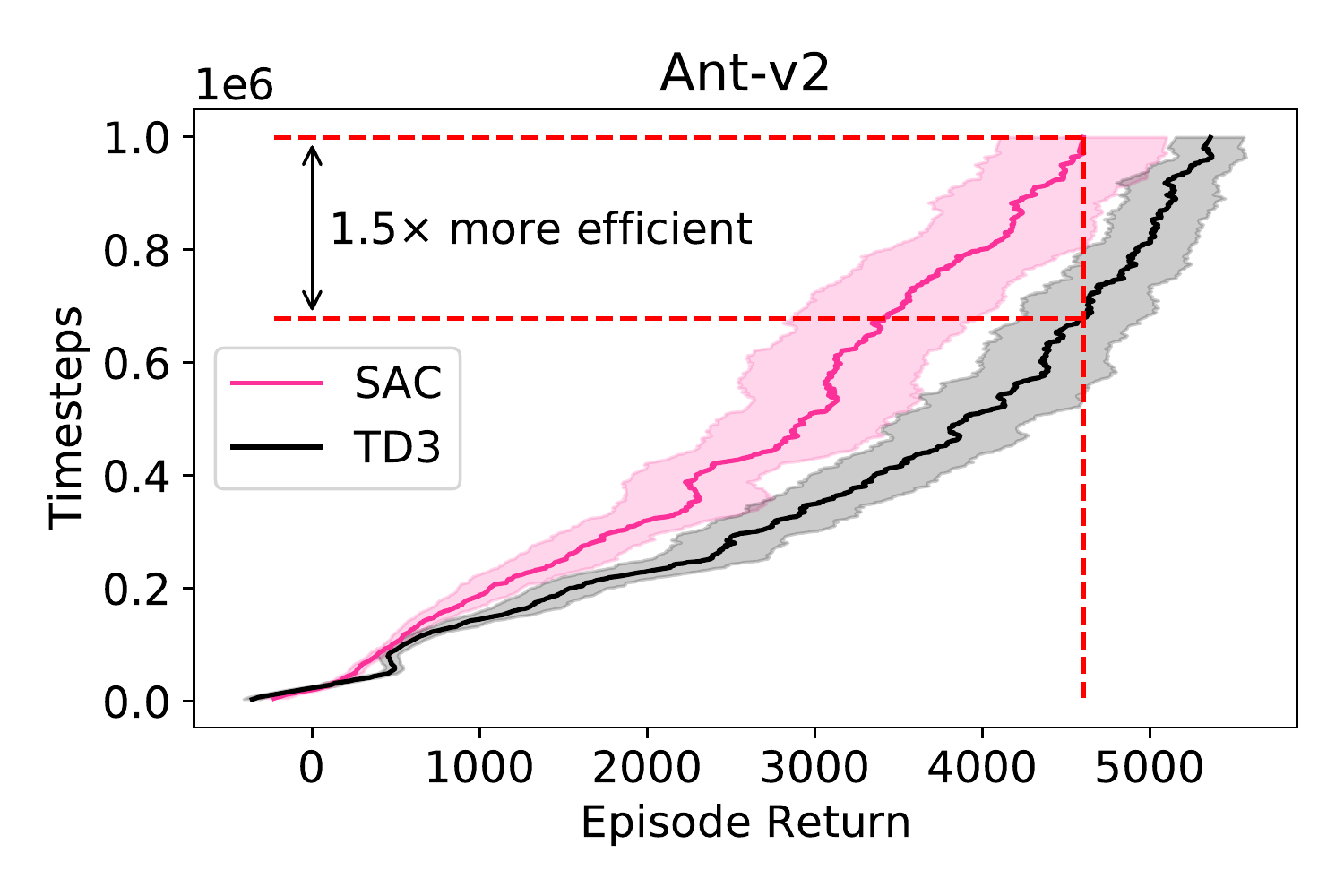}
    \caption{Efficiency gain of TD3 over SAC}
    \label{fig:compareTD3SAC}
\end{figure}

\section{2 v.s. 3 Layers}
\label{app:shallow}

\begin{figure}[H]
\centering
\begin{subfigure}{.45\textwidth}
    \centering
    \includegraphics[width=\textwidth]{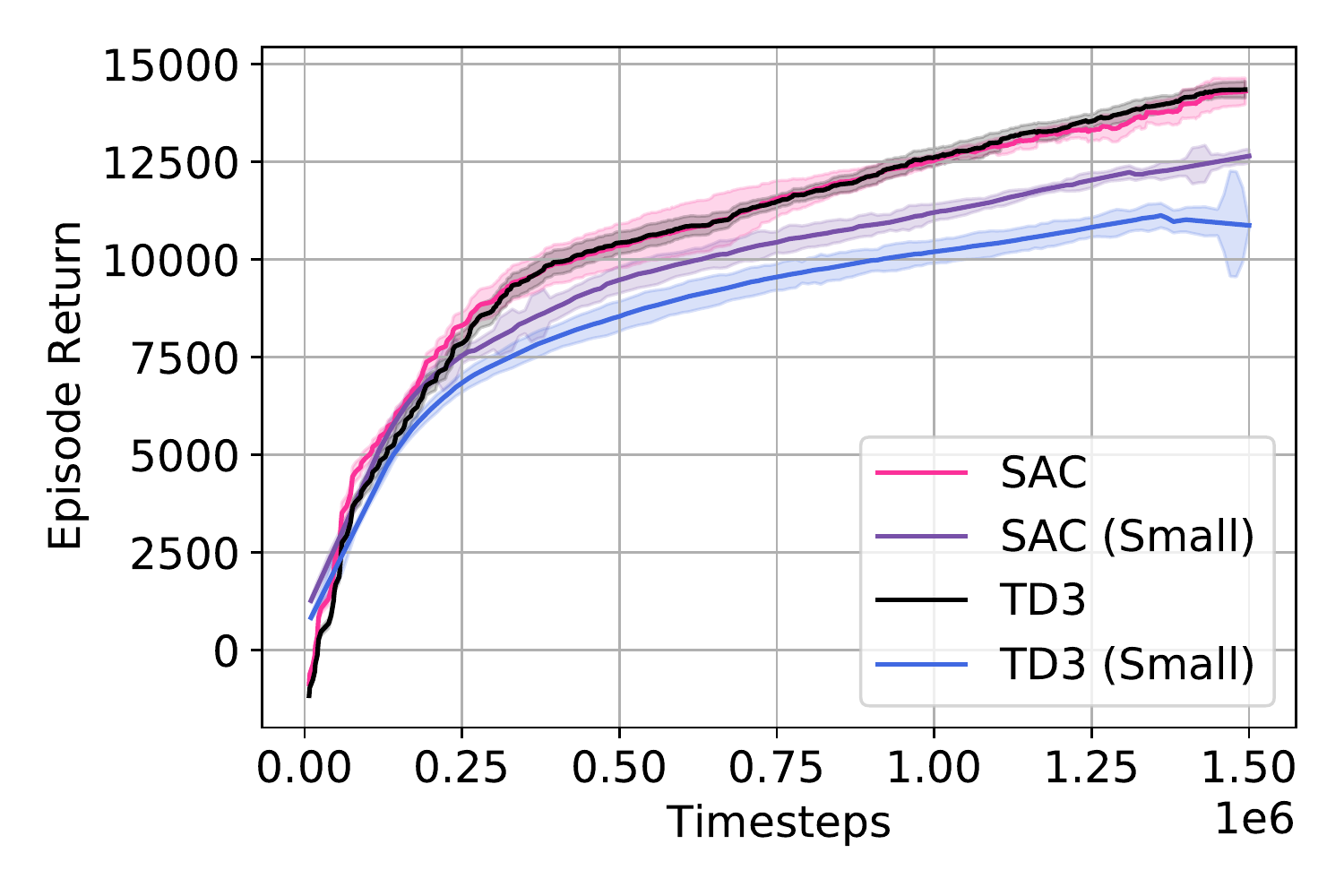}
    \caption{HalfCheetah}
    \label{fig:halfcheetah}
\end{subfigure}
\hspace{0.5cm}
\begin{subfigure}{.45\textwidth}
    \centering
    \includegraphics[width=\textwidth]{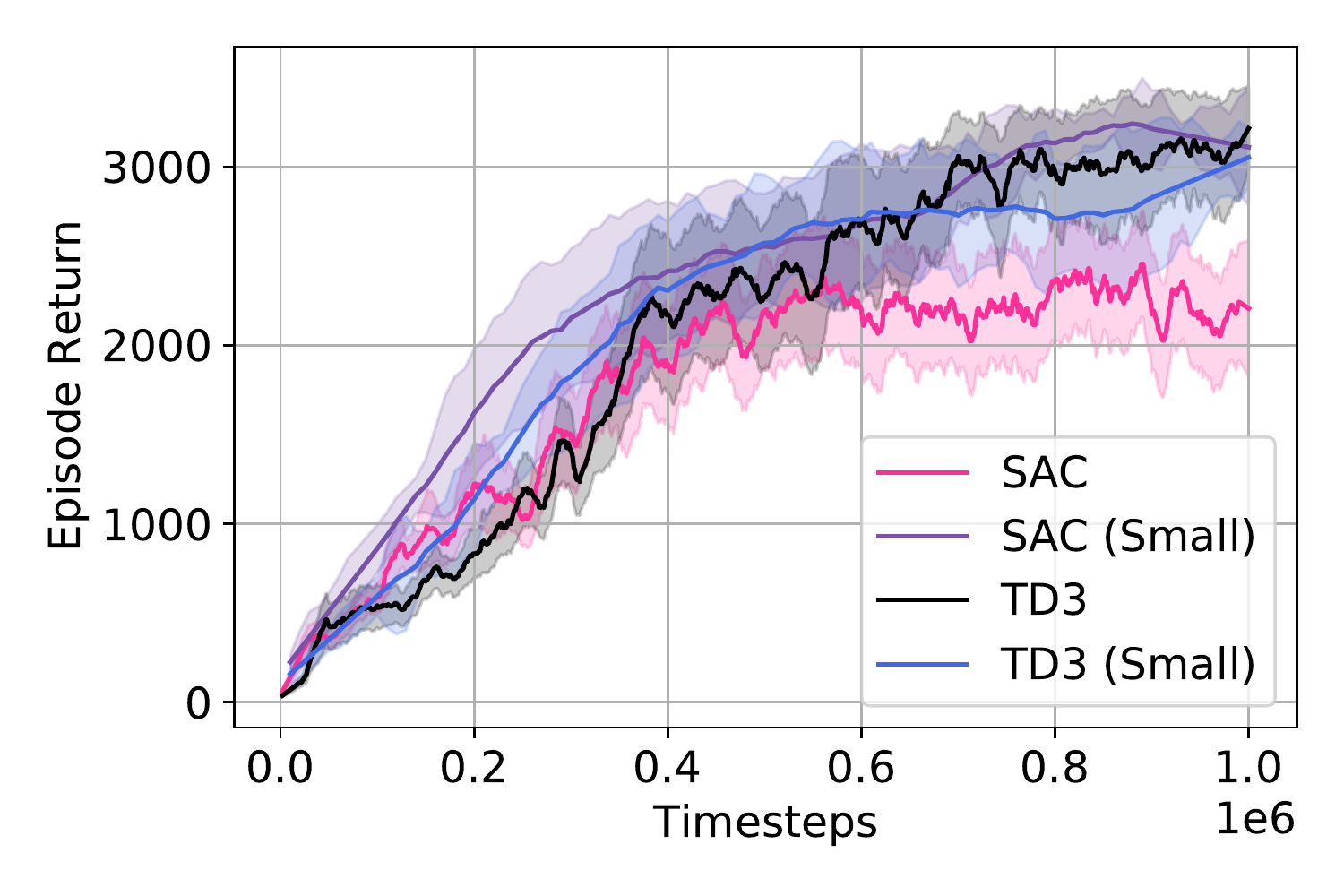}
    \caption{Hopper}
    \label{fig:hopper}
\end{subfigure}
\begin{subfigure}{.45\textwidth}
    \centering
    \includegraphics[width=\textwidth]{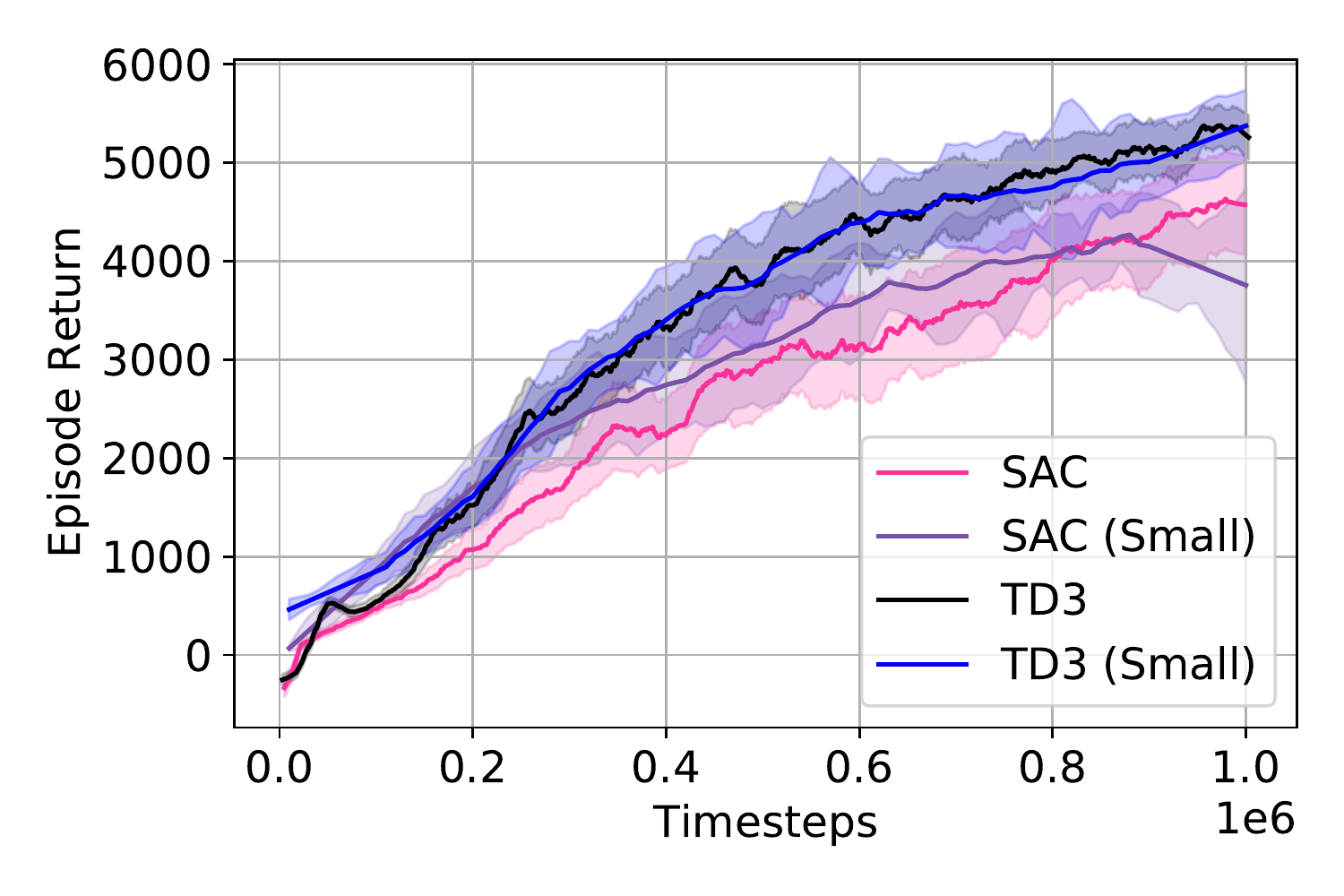}
    \caption{Ant}
    \label{fig:ant}
\end{subfigure}
\begin{subfigure}{.45\textwidth}
    \centering
    \includegraphics[width=\textwidth]{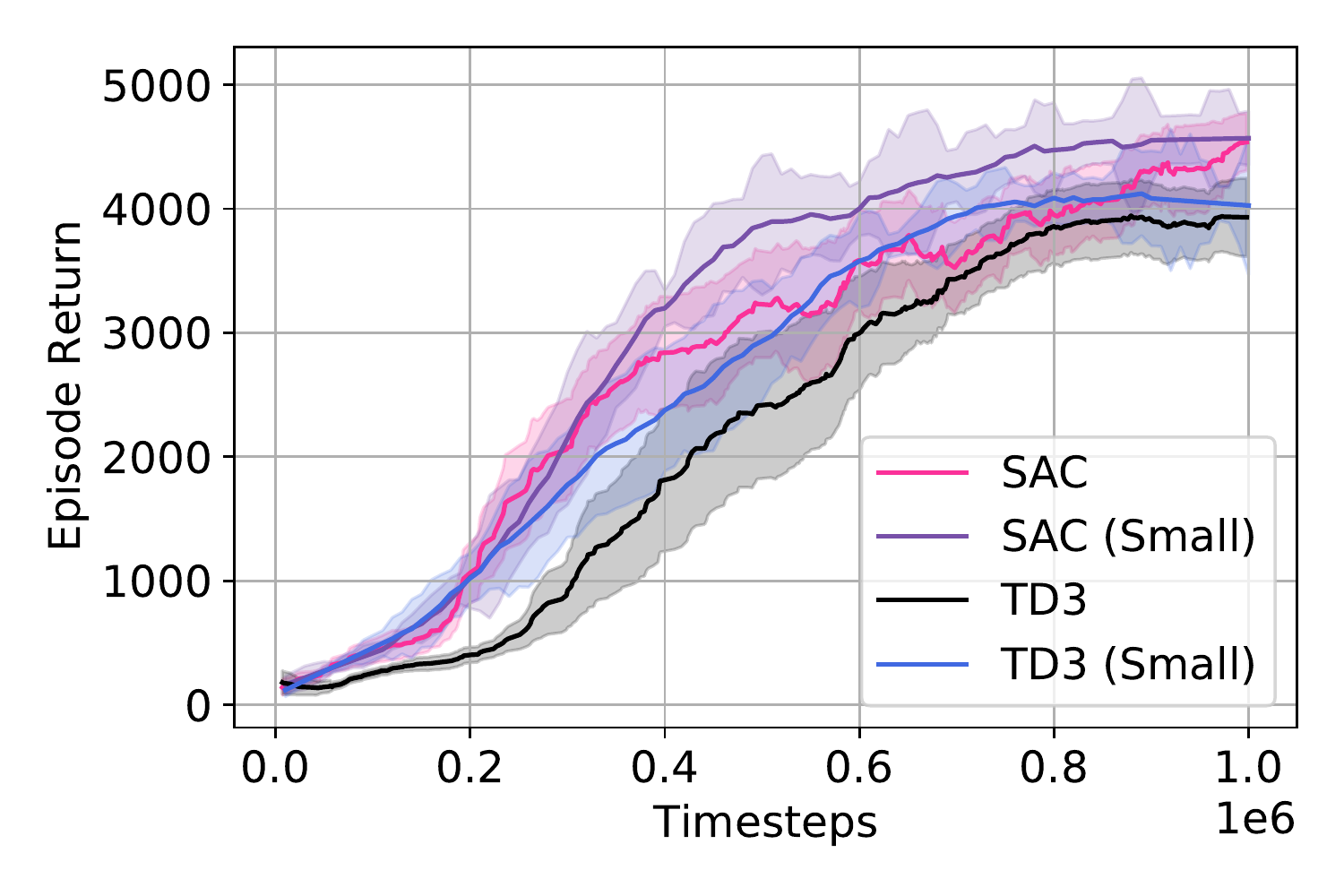}
    \caption{Walker2d}
    \label{fig:walker}
\end{subfigure}
\begin{subfigure}{.45\textwidth}
    \centering
    \includegraphics[width=\textwidth]{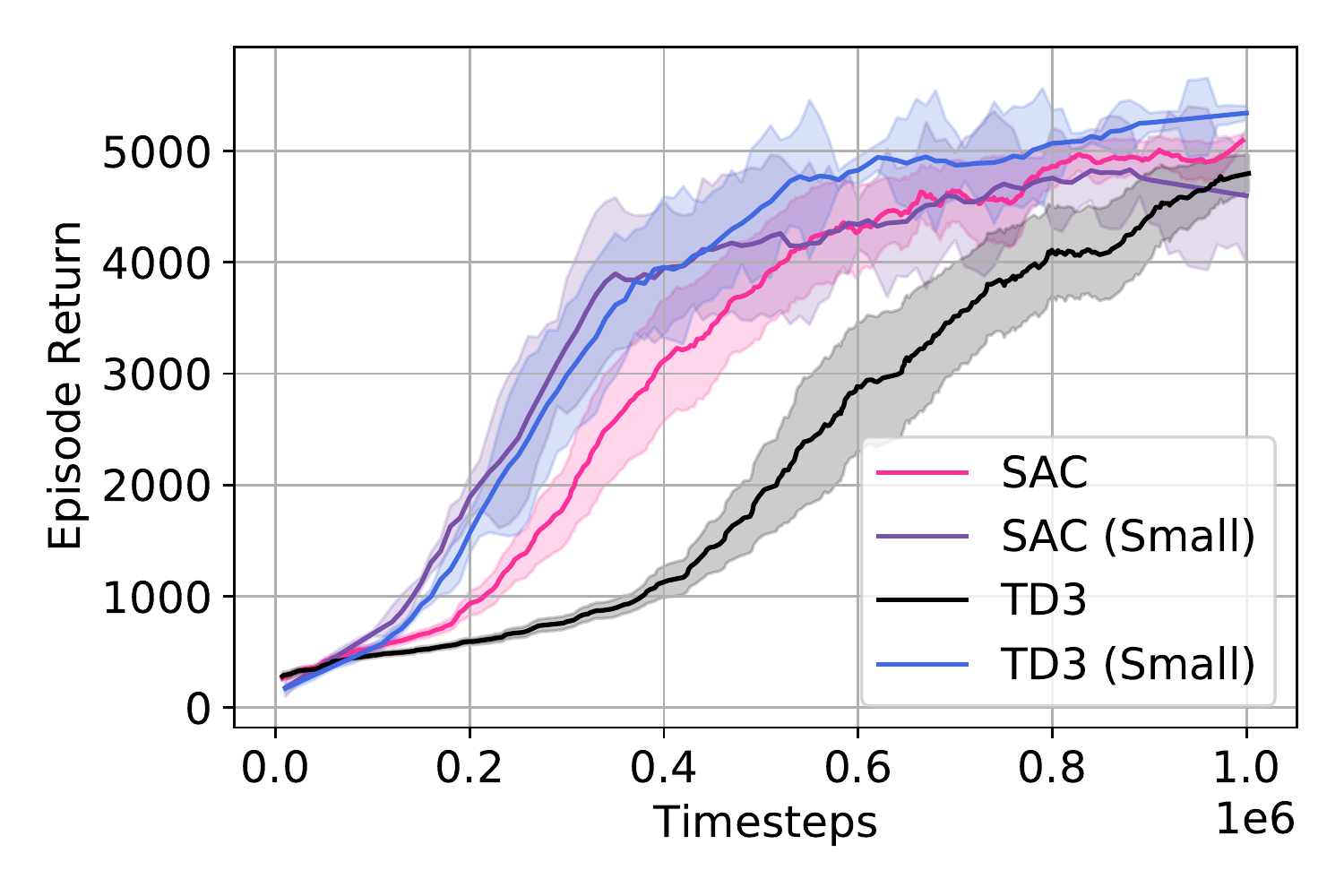}
    \caption{Humanoid}
    \label{fig:walker}
\end{subfigure}
\caption{SAC and TD3 Training Curves on MuJoCo Environments with different network depths (all 5 seeds). (Small) denotes a 2 hidden layer network for both actor and critic.}
\label{fig:compare_shallow}
\end{figure}

Some recent Q-learning work for MuJoCo continuous control has used 3 hidden layers instead of the 2 hidden layers in the original author's code, such as \cite{cql}. Following their lead, and noticing the particularly strong performance on HalfCheetah, we choose to implement 3 hidden layers in \offcon. However, for fairness, we run a set of experiments with 2 hidden layers; the results are displayed in \ref{fig:compare_shallow}. We notice that apart from the significantly improved HalfCheetah performance, and SAC improving on Hopper, the convergent differences are marginal. Note that the small network plots are smoother as we evaluate performance at longer intervals.

\end{appendix}

\end{document}